\documentclass[10pt,twocolumn]{article} 
\usepackage{simpleConference}
\usepackage{times}
\usepackage{graphicx}
\usepackage{amssymb}
\usepackage{url,hyperref}
\usepackage{multirow}
\usepackage{marvosym}
\usepackage{subfig}
\usepackage{booktabs}

\begin{document}

\title{HDFD --- A High Deformation Facial Dynamics Benchmark for Evaluation of Non-Rigid Surface Registration and Classification}

\author{Gareth Andrews$^{1\dag}$, Sam Endean$^{1\dag}$, Roberto Dyke$^2$, Yukun Lai$^2$, Gwenno Ffrancon$^3$, Gary KL Tam$^1$\\
$^1$Department of Computer Science, Swansea University, ($^\dag$equal contribution)\\
$^2$School of Computer Science and Informatics, Cardiff University\\
$^3$Academi Hywel Teifi, Swansea University
}

\maketitle
\thispagestyle{empty}

\begin{abstract}
Objects that undergo non-rigid deformation are common in the real world. A typical and challenging example is the human faces. While various techniques have been developed for deformable shape registration and classification, benchmarks with detailed labels and landmarks suitable for evaluating such techniques are still limited. In this paper, we present a novel facial dynamic dataset HDFD which addresses the gap of existing datasets, including 4D funny faces with substantial non-isometric deformation, and 4D visual-audio faces of spoken phrases in a minority language (Welsh). Both datasets are captured from 21 participants. The sequences are manually landmarked, with the spoken phrases further rated by a Welsh expert for level of fluency. These are useful for quantitative evaluation of both registration and classification tasks. We further develop a methodology to evaluate several recent non-rigid surface registration techniques, using our dynamic sequences as test cases. The study demonstrates the significance and usefulness of our new dataset --- a challenging benchmark dataset for future techniques.
\end{abstract}

\section{Introduction}
We live in a world full of objects that undergo non-rigid deformation. Human faces, animals are some examples. To process scanned deformable objects, techniques are often required to align them, e.g. to reconstruct shapes from individual views, or to provide a canonical basis for statistical analysis. 
Given two surfaces, such registration techniques deform and align the source surface to the target surface.  Non-rigid registration for 3D surfaces is a challenging problem in vision and graphics related areas. Accurate and efficient alignment of surfaces is not only the core of the 3D acquisition and reconstruction pipeline, but also impacts many downstream techniques and applications. Example areas include cross-parametrisation, texture and skeleton transfer, shape interpolation, symmetry detection, object matchings, tracking and statistical shape analysis \cite{tam2013registration}. These further lead to innovative applications like real-time facial puppetry \cite{Weise2009FLF,weise2011realtime,saragih2011real}.

In the past decades, many non-rigid registration techniques have been developed, making different deformation assumptions and therefore capable of handling datasets with different characteristics.
Example datasets include body (e.g. animal, human) deformation \cite{Anguelov2005TOG,Kim2011BIM}, facial movement \cite{Cao14TVCG}, and more recently human body tissues \cite{Bogo14FAUST,Pishchulin2017}. Such datasets allow different registration techniques to be evaluated. Techniques that support articulated (piece-wise rigid) or isometric deformation would work well on large-scale body deformation. To handle more elastic deformation (e.g. facial movement, body tissues), registration techniques that support near to non-isometric deformation are thus required. The recent significant development of robust accurate registration techniques is largely contributed and driven by the availability of public challenging datasets.

With the recent growth and interests in AR/VR techniques, real-time, real-life 3D acquisition techniques have become an important research topic. Further, with the increasing power of capturing devices, 3D dynamics (sequences) will soon become popular. To support capture and registration of long sequences for AR/VR, there is a need for developing more reliable, flexible and accurate registration techniques that work for real-life and challenging examples. These have recently inspired the work of faces in the wild (static 3D faces with free-form deformation) \cite{booth2017itw3dmm}. However, among all the above-mentioned datasets, we observe two main limitations. First, majority of publicly available datasets largely focus on near-isometric (mostly non-elastic) deformation. There is a lack of reliable ground truth datasets that offer real-life non-isometric (highly elastic) deforming sequence data. Second, topology change is a common characteristic of real-life 3D acquisition data due to unavoidable occlusion during 3D capturing process. However, producing such a dataset is difficult due to the tedious annotation efforts involved for long sequences.

Our paper tries to address this research need by offering, as far as we know, the first 3D dynamic (4D) facial dataset with highly deformed funny faces. We choose to capture funny faces -- with highly elastic facial deformation -- because it is frequently observed in daily life (and funny video). In our settings, participants are allowed to pose any facial deformations in an unconstrained manner. The resulting dataset not only offers highly elastic (non-isometric) deformation, but also topological change of mouth (open and close), with or without tongue sticking out. It poses a great challenge to many state-of-the-art techniques as they are not developed to tackle these situations. We also provide ground truth landmarks for this challenging dataset, supporting the development and evaluation of future novel registration techniques. We further develop a methodology for evaluation, using our dataset to provide increasingly challenging test cases. To demonstrate its usefulness, we evaluate our dataset with several state-of-the-art free-form registration techniques.

Apart from the challenging funny faces, to the best of our knowledge, we developed the first visual-audio 4D dataset for the Welsh language. The different consonants and mutations in the Welsh language make it a challenging language to learn. From our survey, most of the visual-audio datasets focus on major speaking languages (e.g. English, French) only; minority languages (like Welsh) are often ignored. 
%This poses a challenging problem 
%to 1) preserve, 2) promote and teach, and 3) evaluate these spoken languages, especially for remote new learners. 
%%%YKL There is nothing wrong with the sentence above, but it is not related to computer vision, so I think we should remove it.
In a controlled setting, participants' audio of different levels of fluency (native to unintelligible) are recorded together with their 3D facial dynamics. This also offers new capacity for researchers to investigate the correlation between 4D facial geometry and pronunciation, and research and evaluation for classification of fluency.
%promoting the teaching and research of Welsh on a global scale, setting the first step for the research of other minority languages.

In summary, our contributions, to the best of our knowledge, include:
\begin{enumerate}
\item The first public 4D funny face dataset offering extreme elastic (non-isometric) deformation, and ground truth annotation of facial feature points. We extended the iBug template \cite{booth2017itw3dmm} to allow not only the facial feature points, but also the tongue to be labelled, across all 20 sequences on all non-static frames, and
%%%YKL Why 10 sequences, surely we have 21 participants?
%%%KLT yes should be 21.
\item The first public 4D visual-audio dataset for Welsh language with different levels of fluency professionally labelled.
\item We develop a methodology for evaluating non-rigid surface registration techniques and apply this to several state-of-the-art registration techniques.
\end{enumerate}

%%%KLT In term of novelty, is it that we are the largest number of annotation points per faces for long sequences? I recall Gareth mentioned something before. Not sure what is the exact wordings. Perhaps you Gareth and Sam can update.
% GDA - We are the first to use iBug to label 4D sequences - only been used for static images previously (James Booth)

The rest of the paper is organised as follows: Section \ref{sect:related} provides a thorough literature survey on static, dynamic and visual-audio 3D dynamic datasets. Section \ref{sect:dataset} discusses the creation of our dataset, from collection to annotation of Welsh fluency and funny faces. Section \ref{sect:evaluate} provides a robust evaluation of our dataset with three state-of-the-art registration techniques, using error metrics and visualisations. Finally, Section \ref{sect:conclude} concludes this paper.

\section{Related Work}
\label{sect:related}
Public datasets are essential to research and development of registration and correspondence techniques. Over the past decades, many datasets have been made available, ranging from static 3D datasets to more recent dynamic 4D sequences. Each dataset targets different kind of non-rigid deformation. We summarise them into three sections. Section \ref{sect:3DGeneral} discusses datasets concerning 3D deformation, mostly with pairwise correspondences. Section \ref{sect:3Dface} discusses datasets focusing on facial sequences. Section \ref{sect:AudioVisual} discusses dynamic datasets with audio features.

\subsection{Datasets for 3D Non-Rigid Surface Registration}
\label{sect:3DGeneral}
Earlier development of public ground truth datasets focused on establishing correspondences between deformed static shapes, notable examples include \cite{Anguelov2005TOG,Bronstein3DOR10,Bronstein2008book,Kim2011BIM,Bogo14FAUST,Pishchulin2017,Cao14TVCG}. \cite{Anguelov2005TOG,Bronstein3DOR10,Bronstein2008book} focus on body shapes (e.g. human, animal) and their correspondences. They are frequently used in registration techniques, assuming mostly piecewise or isometric deformation. Recent work focuses on more challenging large-scale near-isometric body to subtle non-isometric body tissue deformation \cite{Kim2011BIM,Bogo14FAUST,Pishchulin2017} with dense correspondences. However, highly elastic deformation is not common in these datasets.

The 3DFE dataset \cite{yin20063d} targets facial expression recognition. It consists of 100 participants performing seven basic facial expressions (neutral, happiness, disgust, fear, anger, surprise, sadness) at four different levels of intensity. %Therefore there are 25 3D facial expression models for each participant.
\cite{Cao14TVCG} also provides many static human 3D faces focusing on facial expressions. Apart from these, recent research focuses on building 3D facial statistical models on a large collection of 10,000 static faces \cite{james_3DMM}, and facial expressions in the wild \cite{booth2017itw3dmm}. As our focus is to develop a 3D dynamic (4D) facial dataset for challenging elastic deformation, we would refer readers to recent surveys and latest work in \cite{6726093,booth2017largescale}.

%These are examples of many static 3D datasets that have been released to assist the development of facial analysis techniques.

%The authors planned a survey for 4D facial datasets, however at the time of writing it has not been released.

\subsection{Dynamic Facial Datasets}
\label{sect:3Dface}
In this section, we survey publicly available 3D dynamic facial datasets, which provide 3D dynamic sequences data and are more relevant to our work.

The facial action coding system (FACS) \cite{cohn2007observer} was primarily developed by psychologists to describe 44 different facial actions or action units (AUs) that form the basis of 6 facial expressions (happiness, disgust, fear, anger, surprise, sadness). In facial expression datasets, it is common for FACS experts to manually annotate the AUs present in frames.

D3DFACS \cite{6126510} is likely to be the first 3D dynamic dataset for facial expression research, with FACS encoded. The dataset contains 10 participants each performing between 19 and 97 AU sequences for a total of 534 sequences. Each sequence is 5 to 10 seconds long. Participants were asked to perform a particular AU repeatedly. The performance closest to the AU was extracted from onset to peak and scored by a FACS expert.

BP4D \cite{6553788} is a 3D dynamic facial expression database and addresses the need for spontaneous facial expressions. Spontaneous facial expressions are more natural expressions obtained from sudden impulse, without any external stimulus. 41 participants were subjected to various activities to elicit the expressions of 8 different emotions. Activities included experiencing an unpleasant smell (disgust) and submerging their hand into ice water (physical pain). The dataset also includes 20 seconds of manually annotated AU’s (FACS) where the facial expression was most productive and 2D/3D facial landmarks for each sequence. A Di3D system was used to capture the sequences at 25fps.

BP4D+ \cite{7780743} is a multi-modal dynamic dataset incorporating 2D and 3D facial dynamics, skin temperature dynamics and physiological responses e.g. blood pressure, heart rate and respiration. 140 participants engaged in a range of activities to elicit spontaneous emotions, such as, watching videos, experiencing pain and talking with professional actors. Expert FACS coders annotated AU’s for 4 sequences for all participants. The dataset was captured by a Di3D System.

UMB-DB \cite{6130509} is a 3D dataset of 1473 facial scans of various expressions, of which 578 scans have been partially occluded by various objects such as hats, scarves and hands. The dataset is made up of 9 acquisitions of 143 participants (98 male / 45 female): 3 with a neutral expression, 3 with non-neutral (smiling, angry, bored) and 3 with a partially occluded face. Data was captured using a Minolta Vivid 900 laser scanner.

These datasets have been very useful for detection, recognition, synthesis of faces and facial expressions. However, none of these datasets contain extreme elastic deformation of faces which is the focus of this paper.

\subsection{Visual-Audio Datasets}
\label{sect:AudioVisual}
This section discusses relevant works that integrate speech into 3/4D datasets.

IV\textsuperscript{2} is a multi-modal dataset that includes 300 participants each speaking 15 French sentences \cite{unknown}. However these sequences were only captured in 2D. Moreover the dataset also contains iris data, 2D stereoscopic images and static 3D face scans for each participant, captured using a Minolta Vivid 7000 scanner.

In the XM2VTSDB \cite{messer1999xm2vtsdb} and BMDM \cite{4815263} datasets, participants were instructed to enunciate digits from “0” to “9” and some spoken phrases. The intended uses for these datasets are speech synthesis and lip reading. However these databases are both 2D, 
with no 3/4D data available.
%and no 3/4D dataset is available.

Cosker et al. released a 4D dataset of non-verbal facial actions \cite{5438780}. A number of expressions were chosen that require activation of various muscle regions. 94 participants were each asked to utter the word `puppy' and a further 15 of which were asked to utter other words e.g. `password', `mushroom'.

2D CCDb \cite{Aubrey_2013_CVPR_Workshops} is a 2D visual-audio dataset containing 30 natural conversations between pairs of people where two synchronised 3dMD systems were used during the capturing process. In subsequent work, a 3D dynamic dataset, 4D CCDb, was released \cite{marshall20154d} which contains 17 minutes of natural conversations, each captured at 60fps. 4 participants were each captured engaging in conversation with the others. A variety of facial expressions and head motion were annotated and released with the dataset.

Despite the great research effort to provide FACS annotation, all these datasets  mainly target emotion and speech recognition. To the best of our knowledge, none of these datasets provide facial landmarks for each frames of the sequences. Therefore, they cannot be used as ground truth for evaluation of non-rigid surface registration techniques.

In our literature search, visual-audio datasets are useful for emotion and speech recognition. There are many 2D audio-visual datasets released in past decades, and can be found in a recent survey paper \cite{wu_lin_wei_2014}. However, datasets covering 4D sequences of spoken phrases are limited.

With regard to the Welsh language, an attempt is currently being carried out to capture the largest spoken Welsh corpus with the support of a downloadable app \cite{knight17baal}. The corpus includes both video and audio data. However, we are not aware of any 3D dynamic sequence of spoken Welsh phrases. To address this visual-audio 4D challenge, we capture sequences of Welsh spoken phrases with varying levels of fluency. Sequences are manually annotated with facial landmarks, which can be served as ground truth of registration techniques. The dataset also provides expert labelling of fluency, which will be useful for evaluating 4D classification techniques.
%The dataset also serves as a preservation of such a minority language at this contemporary time.

\section{Our Dataset}
\label{sect:dataset}
As made evident above, there exist many datasets covering facial expressions, but few 4D visual-audio datasets with speech incorporated. Further, no dataset provides an extreme elastic deformation of the faces with asymmetric deformation (e.g. funny faces), and topological changes (e.g. mouth open/close, tongue sticking out). In this work, we present a new facial 3D dynamic dataset called HDFD.
Our dataset consists of sequences of 21 participants, each enunciating 10 Welsh phrases. We further provide a Funny face dataset where a challenging face sequence is produced by each participant. We have provided manually placed landmark points on 40 sequences of the dataset using the extended iBug68 template \cite{ibug} with 73 landmark points.

In the followings, we first discuss the procedure for data collection in Section~\ref{sect:collection}. We then discuss the property and annotation of the Welsh dataset in Section~\ref{sect:WelshProperty}, and then the property and annotation of the funny dataset in Section~\ref{sect:FunnyProperty}.

\subsection{Procedure for Data Collection}
\label{sect:collection}

To produce the proposed 4D facial dataset, we recruited 21 volunteers from university staff and students, and compensated their time (around one hour) with stipends. The participants' ability to converse Welsh varied from no experience (never heard of or use Welsh before) to native (Welsh first language, very fluent).

Prior to any capturing, a Welsh expert chose ten Welsh phrases with varying speaking difficulties. To assist non-fluent speakers, a Welsh native speaker was recorded speaking each of the ten phrases in a normal video recording.

During the capturing session, each participant was asked if they would like to be shown the video recording from the Welsh native speaker. Participants are allowed to watch the video any number of times (at least two). 4D recording of the participant would only proceed if they felt ready. Typically, each phrase was recorded for approximately four seconds; however, in cases where participants struggled with a phrase, the recording would be extended to five or six seconds.

%%%RMD there is some ambiguity in the grammar here, "with and/or without their mouth open, and/or with and/or without their tongue sticking out". It is not clear whether the dataset contains 1, 2 or 4 funny face captures.
To conclude the capturing session, participants were then asked to pose one highly deformed, unconstrained, asymmetric funny face. The participants may have their mouth open or their tongues sticking out, but it is their choices. They were asked to start from their neutral face, slowly progress to the peak of their funny face in three seconds, and hold their funny face for 1-2 seconds.

All 4D recordings were captured using a 3dMD system. Each sequence was shot at 48 fps, with audio synchronised and captured simultaneously. The sensing technology of the 3dMD system is based on infra-red and is able to reconstruct 3D surfaces through its 6 stereo cameras (four infra sensors for geometry reconstruction, and two colour sensors for facial texture). Each image amounts to roughly 5MB of bitmaps, and therefore, the system processes around 1.4 GB of raw data per second. This data is stored locally on the system hard drive and subsequently processed overnight to recover the 3D geometry and texture.

%The 20 participants were each recorded speaking, to the best of their ability, 10 Welsh phrases of increasing complexity. The participant's ability to converse fluently in Welsh varied from no experience to fluent (native, Welsh first language).

%Prior to any capturing a Welsh expert was recorded speaking each of the ten phrases to assist non-fluent speakers. For each of the ten phrases, the participant would be shown the video of the phrase being enunciated a minimum of two times, only once they were ready would we record them speaking. Typically each phrase would be recorded for four seconds, but in some cases where participants struggled with the phrase, this would be extended to five or six seconds. To conclude the capturing session, the participant would be recorded performing a highly deformed, asymmetric funny face of their choice. They would be asked to start from their resting face and over 3 seconds, slowly progress to the funny face and hold for 1-2 seconds.

%The dataset was captured using a temporal-3dMD System. Each sequence was shot as 48 fps and is roughly 4-5 seconds long. More information on the data can be found in table... % Statistics on sequences (vertices, faces, gb etc) 

\subsection{Dataset Property and Annotation - Welsh}
\label{sect:WelshProperty}

\begin{table*}[b]
\centering
\caption{Welsh phrases, with increasing difficulties}
\label{tbl:WelshPhrases}
\begin{tabular}{@{}lll@{}}
\toprule
Phrase & Welsh                    & Meaning in English            \\ \midrule
V1     & Eisteddfod yr Urdd       & Welsh Youth Music Competition \\
V2     & Prynhawn da bawb         & Good afternoon everyone       \\
V3     & Dyn busnes yw e          & It's a businessman            \\
V4     & Papur a phensil          & Paper and pencil              \\
V5     & Ardderchog               & Excellent / Superb            \\
V6     & Llwyddiant ysgubol       & Great success                 \\
V7     & Yng nghanol y dref       & In the town center            \\
V8     & Dwy neuadd gymunedol     & Two community halls           \\
V9     & Llunio rhestr fer        & Shortlisted                   \\
V10    & Gwybodaeth angenrheidiol & Necessary information        
\end{tabular}
%\vspace{-0.5cm}
\end{table*}

\begin{table*}[t]
  \centering
  \caption{Welsh fluency of each participant.}
  \label{tbl:WelshFluency}
\begin{tabular}{l|cccccccccc|c}
\textbf{Participant} & \textbf{V1} & \textbf{V2} & \textbf{V3} & \textbf{V4} & \textbf{V5} & \textbf{V6} & \textbf{V7} & \textbf{V8} & \textbf{V9} & \textbf{V10} & \multicolumn{1}{l}{\textbf{average}} \\
\hline
\multicolumn{1}{c|}{\textbf{A}} & 5     & 5     & 5     & 5     & 5     & 5     & 5     & 5     & 5     & 5     & 5 \\
\multicolumn{1}{c|}{\textbf{B}} & 5     & N/A   & 5     & 5     & N/A   & 5     & 5     & 5     & 5     & N/A   & 5 \\
\multicolumn{1}{c|}{\textbf{C}} & 4     & 5     & 4     & 5     & 5     & 5     & 5     & 5     & 4     & 5     & 4.7 \\
\multicolumn{1}{c|}{\textbf{D}} & 2     & 3     & 3     & 3     & 4     & 2     & 4     & 2     & 3     & 3     & 2.9 \\
\multicolumn{1}{c|}{\textbf{E}} & \dag  & 5     & 4     & 5     & 5     & 4\dag & 4     &\dag   & 3     & 3     & 4.14 \\
\multicolumn{1}{c|}{\textbf{F}} & 2     & 4     & 4     & 3     & 4     & 5     & 4     & 2     & 3     & 2     & 3.3 \\
\multicolumn{1}{c|}{\textbf{G}} & 1     & 1     & 2     & 1     & 3     & 1     & 3     & 1     & 3     & 4     & 2 \\
\multicolumn{1}{c|}{\textbf{H}} & 4     & 5     & 5     & 5     & 5     & 5     & 5     & 3*    & 5     & 2     & 4.56 \\
\multicolumn{1}{c|}{\textbf{I}} & 2     & 4     & 2*    & 1     & 3     & 3     & 4     & 3     & 4     & 1     & 2.78 \\
\multicolumn{1}{c|}{\textbf{J}} & 1     & 2     & 3     & 4     & 4     & 4     & 3     & 3     & 4     & 4     & 3.2 \\
\multicolumn{1}{c|}{\textbf{K}} & 4     & 5     & 5     & 5     & 5     & 4     & 5     & 4     & 4     & 5     & 4.6 \\
\multicolumn{1}{c|}{\textbf{L}} & 3     & 2     & 3     & 4     & 4     & 2     & 3     & 2     & 3     & 2     & 2.8 \\
\multicolumn{1}{c|}{\textbf{M}} & 1     & 4     & 3     & 3     & 3     & 3     & 2     & 3     & 3     & 4     & 2.9 \\
\multicolumn{1}{c|}{\textbf{N}} & 5     & 5     & 5     & 5     & 5     & 4     & 5     & 5     & 5     & 5     & 4.9 \\
\multicolumn{1}{c|}{\textbf{O}} & 5     & 5     & 5     & 5     & 5     & 5     & 5     & 5     & 5     & 5     & 5 \\
\multicolumn{1}{c|}{\textbf{P}} & 4     & 4     & 4     & 5     & 3     & 4     & 3     & 3     & 4     & 3     & 3.7 \\
\multicolumn{1}{c|}{\textbf{Q}} & 3     & 3     & 3     & 4\dag & 4     & 4     & 3     & 1     & 4     & 1     & 2.89 \\
\multicolumn{1}{c|}{\textbf{R}} & 1     & 1     & 2     & 2     & 1     & 0     & 2     & 0     & 1     & 1     & 1.1 \\
\multicolumn{1}{c|}{\textbf{S}} & 5     & 5     & 4     & 5     & 5     & 5     & 5     & 5     & 5     & 5     & 4.9 \\
\multicolumn{1}{c|}{\textbf{T}} & 3     & 0     & 2     & 2     & 3     & 2     & 1     & 3\dag & 1     & 2     & 1.78 \\
\multicolumn{1}{c|}{\textbf{U}} & 5     & 5     & 5     & 5     & 5     & 5     & 5     & 5     & 5     & 5     & 5 \\
%\hline
%\textbf{average} & 3.25  & 3.65  & 3.8   & 3.9   & 4.05  & 3.65  & 3.86  & 3.28  & 3.76  & 3.35  &  \\
\end{tabular}%

\begin{flushleft}
\scriptsize
%\footnotesize
N/A: Video \& Audio is unavailable, \dag: Audio is incomplete, *Read incorrectly\\
5 = Totally fluent / No mistakes   \\
4 = Fluent but 1 or 2 mistakes / Slight non-Welsh accent in parts\\
3 = More than 2 mistakes / Non-Welsh accent           \\
2 = Intelligible but many mistakes / Heavy non-Welsh accent\\
1 = Only just intelligible / Hard to understand\\
0 = Unintelligible / Cannot understand at all
\end{flushleft}
\vspace{-1cm}
%\vspace{-0.5cm}
\end{table*}

Our dataset contains Welsh phrases from 21 participants. Welsh is a minority language and is less frequently researched. Commercial and automatic speech recognition systems for Welsh are mostly unavailable. Welsh was further chosen due to its complex phonemes and the resultant difficulty in enunciation. This may result in elaborate facial movements, which may lead to challenges and complications for automatic speech recognition and analysis.

The ten phrases (Table \ref{tbl:WelshPhrases}) were chosen by a Welsh expert (a professional development and quality manager from the Academi Welsh for Adults, who is also one of the most experienced tutors and Welsh language experts in the University). These Welsh phrases would reveal the common problems of Welsh learners, and can be used to gauge the level of Welsh fluency of the speakers.

After all capturing sessions, the captured videos are played back to the Welsh expert for annotation of level of fluency. The Welsh expert defined six levels of Welsh fluency, ranking from 0 to 5, with 0 being unintelligible and 5 being totally fluent. The fluency annotation is made available in Table \ref{tbl:WelshFluency}. The Welsh expert also considered participants \textbf{H} and \textbf{I} voiced out the wrong Welsh words, due to some subtle difference the participants may not recognised. These are marked as ``*Read incorrectly''. Table \ref{tbl:WelshFluency} also shows that there is a wide range of Welsh fluency in our dataset, with averages ranging from 1.1 to 5.

All data was captured with a 3dMD facial dynamic capture system. Due to the large volume of data to be handled by the 3dMD system (1.4GB/s), the system occasionally failed to capture any data due to bandwidth issues and limited cache size. Two affected participants were then invited to recapture the phrases if they were available. Overnight 3D reconstruction is also a time-consuming step that we cannot immediately verify the successful reconstruction before participants left. Because of these practical issues and second invitation, three sequences (participant \textbf{B}) are declared lost. Sometimes, 3dMD system occasionally chopped at the beginning leading to incomplete audio streams (participants \textbf{E}, \textbf{Q} and \textbf{T} for certain Welsh phrases). The Welsh expert still tried to help us annotate the audio recording.

Despite all these issues, we have successfully captured all ten Welsh phrases spoken by 17 participants (at least seven Welsh phrases for all 21 participants). Each phrase was captured for 4-5 seconds, which equates to around 200-240 frames. A single frame consists of mesh data ($\sim$17K vertices) and a texture bitmap ($\sim$5MB) per file. Each sequence takes approximately 1.8GB.

%With the increasing difficulties of Welsh phrases (V6-V10), we can observe that majority of non-experienced Welsh speakers fail to produce high level of fluency. In general, a good the spectrum of Welsh fluency (fluent, medium to unintelligible) is available in our dataset.

Moreover we provide proper landmark annotation to phrases \textbf{V2} \& \textbf{V10}. These phrases were selected due to their contrasting difficulties. Landmarking was facilitated by the Landmarker tool freely offered by the Menpo project \cite{menpo14}. The landmarking was an extensive process (roughly 1 hour per sequence) that was cross validated by another person to ensure precision and accuracy for the subsequent evaluation. Consequently, annotation is provided for ten participants, on frames where the participant is speaking only (approximately 50 - 150 frames). 

\begin{figure*}[t]
\centering
\subfloat[native]{\includegraphics[width=0.49\linewidth]{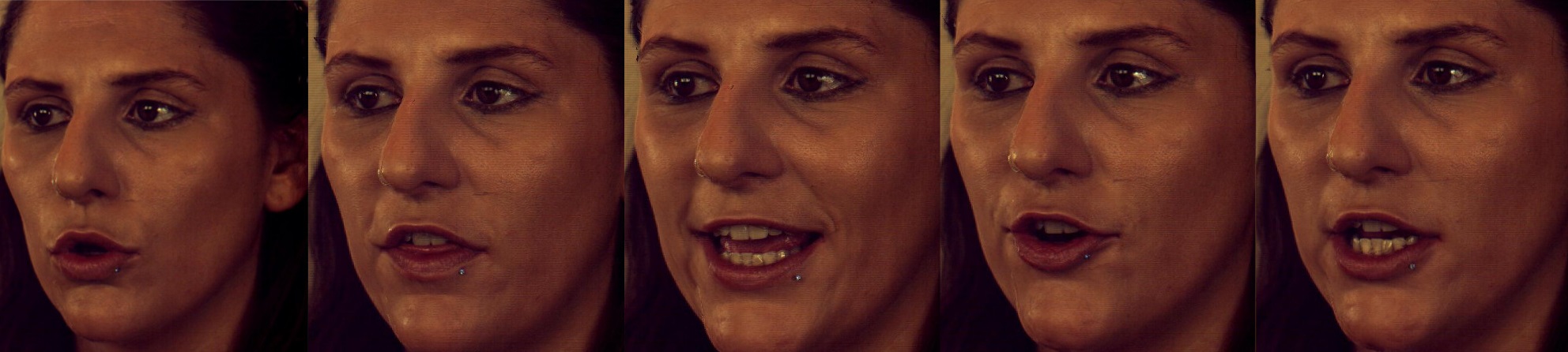}\label{fig:nativeA}}\hspace{0.02cm}
\subfloat[native]{\includegraphics[width=0.49\linewidth]{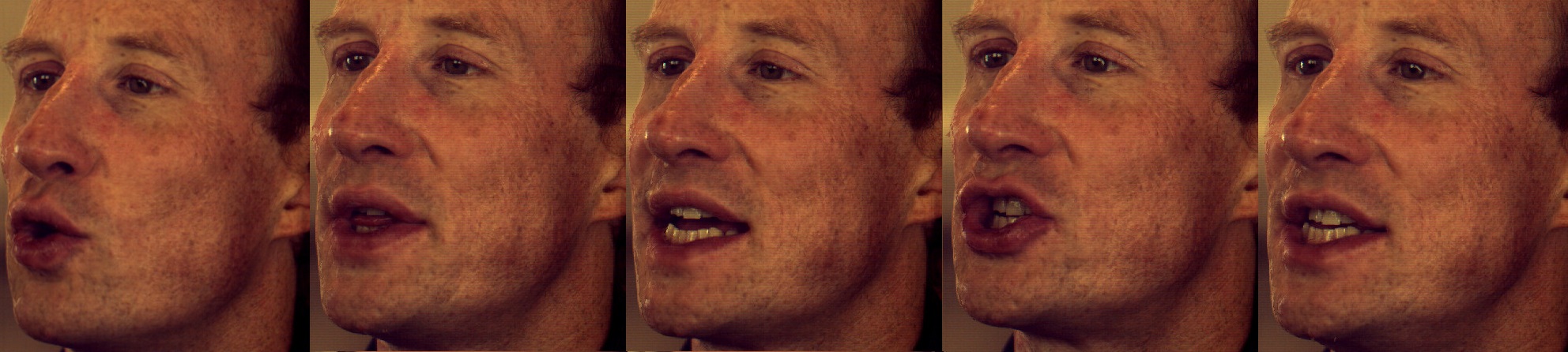}\label{fig:nativeB}}\\
\subfloat[high performing non-native]{\includegraphics[width=0.49\linewidth]{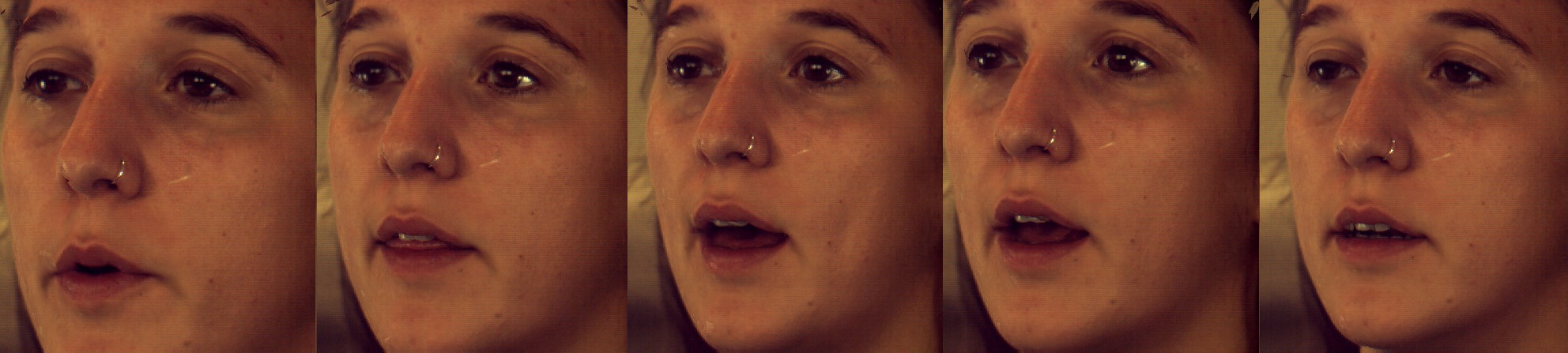}\label{fig:nonnativeC}}\hspace{0.02cm}
\subfloat[low performing non-native]{\includegraphics[width=0.49\linewidth]{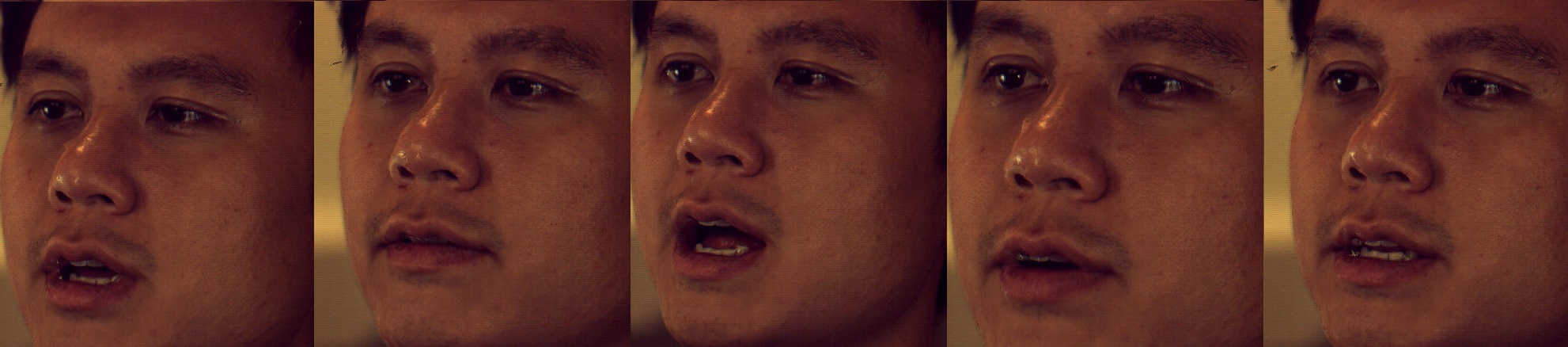}\label{fig:nonnativeD}}
\caption{\label{fig:seq10analysis}The lips and mouth deformation of four participants pronouncing \textbf{V10} (Gwybodaeth angenrheidiol). \protect\subref{fig:nativeA} and \protect\subref{fig:nativeB} are two Welsh fluent participants. \protect\subref{fig:nonnativeC} is a high performing non-native, and \protect\subref{fig:nonnativeD} is a low performing non-native. These frames highlight the participant's mouth at certain phonemes for comparison. The protrusion of the lips in Frames 1 and 4, and the extent of the ``ee'' phoneme in Frame 3 is much greater in fluent speakers. The use of the tongue in the ``eth'' sound of Frame 2 is not evident in the low performing participant. This suggests that a strong correlation between the 3D shapes of mouths/lips and fluency may exist, facilitating the analysis of learning and analysis of Welsh. (Readers are refer to Section \ref{sect:evaluate} for the meaning of Frames 1-5.)}
\vspace{-0.5cm}
\end{figure*}

The Welsh speaking 4D dataset is the first of its kind, allowing preservation of speaking Welsh at this contemporary time. From our earlier observation, 3D dynamics may help better analyse how Welsh is natively spoken (Figure \ref{fig:seq10analysis} and its caption description). With the annotation of fluency and facial landmarks, the dataset can be served as training set for advanced machine learning models, teaching, evaluation, and the system development of speech recognition, classification, synthesis and lip reading purpose for Welsh.

%We further provided landmark annotation of such Welsh sequences on participant faces.

\subsection{Dataset Property and Annotation - Funny Faces}
\label{sect:FunnyProperty}
After capturing the Welsh phrases, each participant is asked to pose a challenging funny face. This is the second part of our dataset, which features highly elastic, unconstrained, asymmetrical facial deformation. It presents a real challenge for registration methods in two ways. First, the faces undergo extreme non-isometric (elastic) changes. To handle such datasets, a deformation model with high degree of freedom is usually required, but would easily lead to overfitting problem \cite{tam2013registration}. The peak frame of each funny face sequence can be seen in figure \label{fig:funnybanner}. Second, users are unconstrained to open their mouths or have their tongues sticking out. When the tongue sticks out, the elongation of the tongue is highly elastic (non-isometric). Opening mouth would also create a hole in the face, leading to change in topology on the underlying 3D surfaces. Both of these would impact many well-known near- to non-isometric deformation techniques e.g. \cite{HUANG2008,Kim2011BIM} which assume sphere topology or use of geodesic distances to approximate isometry. 

% \begin{figure}
%     \centering
%     \begin{minipage}{0.4\textwidth}
%         \centering
%         \includegraphics[width=1\textwidth]{template.jpg}
%         \label{fig:template}
%         \caption{Our extended iBug template}
%     \end{minipage}\hfill
%     \begin{minipage}{0.3\textwidth}
%         \centering
%         \includegraphics[width=1\textwidth]{Marceli_high_def.PNG}
%         \label{fig:assymetric}
%         \caption{Funny face with extreme and asymmetrical mouth deformation}
%     \end{minipage}\hfill
%     \begin{minipage}{0.3\textwidth}
%         \centering
%         \includegraphics[width=1\textwidth]{Sion_w_lm.PNG}
%         \label{fig:tongue}
%         \caption{Funny face with tongue sticking out}
%     \end{minipage}
% \end{figure}

\begin{figure*}[t]
\centering
\includegraphics[width=\textwidth]{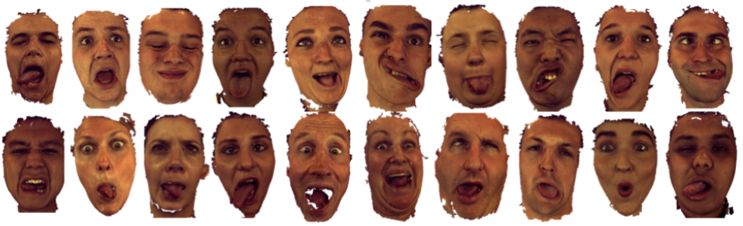}
\vspace{-0.2cm}
\caption{The peak frame of each funny face sequence we have captured. }
\label{fig:funnybanner}
\vspace{-0.6cm}
\end{figure*}

\begin{figure*}[t]
\centering
%\subfloat[]{\includegraphics[width=0.38\textwidth]{LandmarkPoints.png}\label{fig:ibug73}}
%\subfloat[]{\includegraphics[width=0.33\textwidth]{1.png}\label{fig:assymetric}}
%\subfloat[]{\includegraphics[width=0.29\textwidth]{2.png}\label{fig:tongue}}
\subfloat[]{\includegraphics[height=4cm]{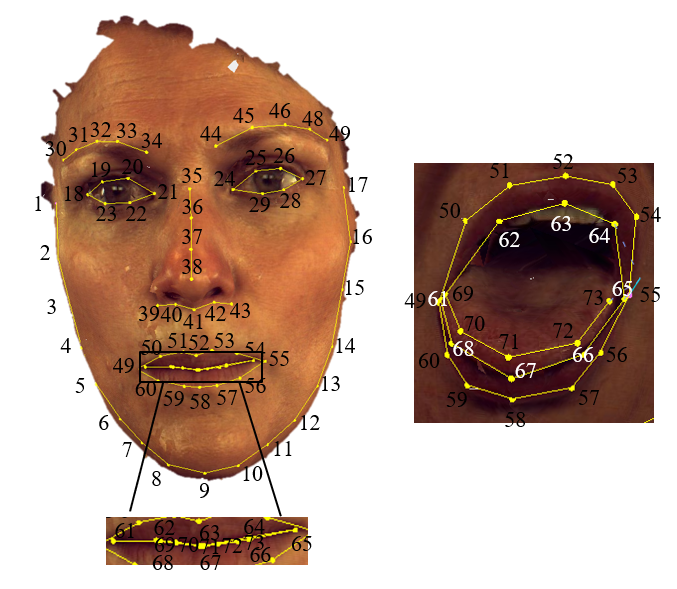}\label{fig:ibug73}}
\subfloat[]{\includegraphics[height=4cm]{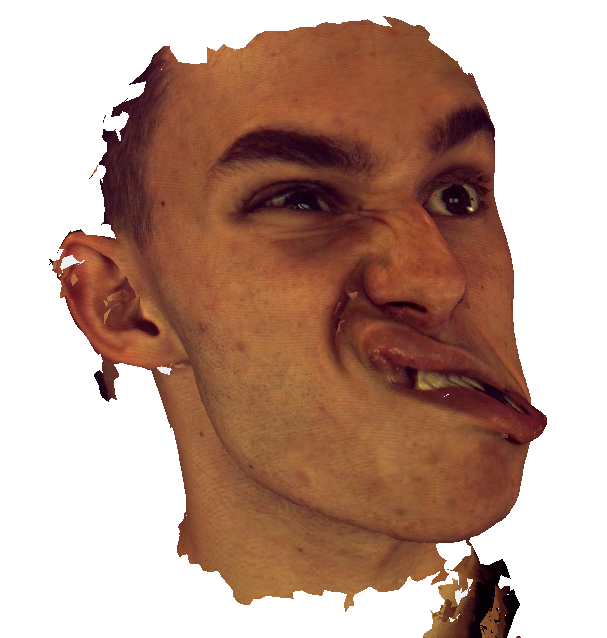}\label{fig:assymetric}}
\subfloat[]{\includegraphics[height=4cm]{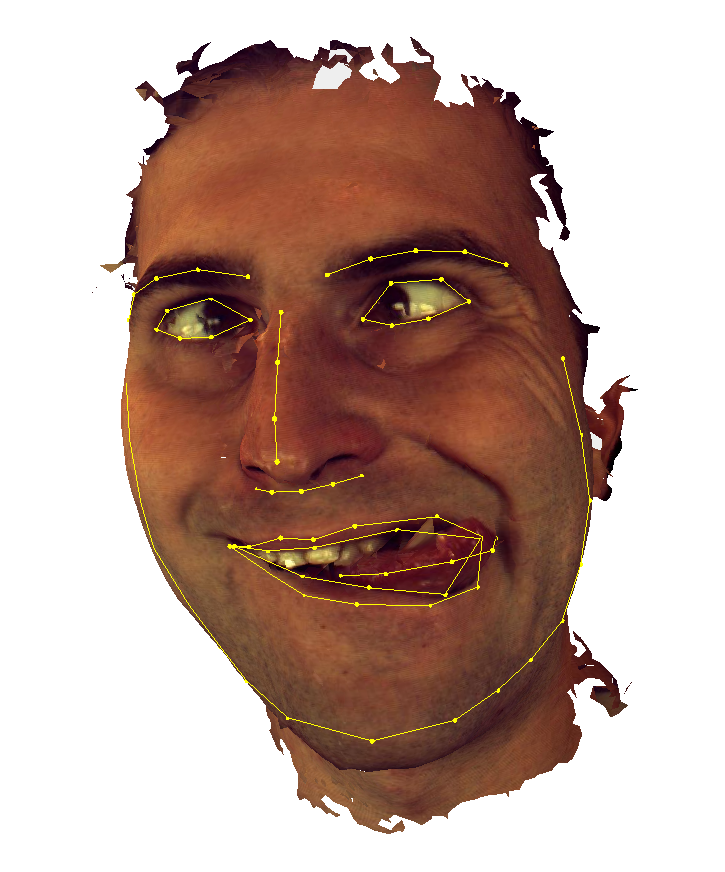}\label{fig:tongue}}
\caption{\label{fig:template}Our extended iBug template and example highly deformed faces. \protect\subref{fig:ibug73} template face with 73 landmark points. Example annotations of tongues are also shown. \protect\subref{fig:assymetric} Funny face with extreme and asymmetrical mouth deformation. \protect\subref{fig:tongue} Funny face with tongue sticking out.}
%\vspace{-0.4cm}
\end{figure*}

%In particular, it is the movement of the mouth and the tongue that present significant challenge, as they result in uncommon changes to topology. When the mouth opens.
%The occlusion caused by the tongue also creates issues for the stereo imaging system itself.
%%%KLT I have not idea why tongue would affect the stability of the imaging system.
%%% extension of tongue is also non-isometric deformation
%%%KLT elaborate... what does it mean? Reviewer will ask: does it mean your system is not reliable? Is the dataset trustable?

%%%KLT Give statistics, is it 40000 3D models? How do you get the number? Tell it here.

In our dataset, we have captured a funny face sequence each for 20 participants (Figure \ref{fig:funnybanner}). To landmark these deformation sequences, we used the landmarker tool \cite{menpo14}. In our settings, it was necessary to extend the iBug68 template \cite{ibug} to include five additional landmark points featuring the movement of the tongue (see Figure \ref{fig:template}). The landmarking process thus consisted of manually placing 73 points on the surface of each 3D model to identify the key features of the face. The labelling process takes around 4 hours per sequence, and was carried out by two persons. Upon completion, the landmarks were cross validated to ensure precision and consistency for the subsequent evaluation.

% \begin{figure}
% 	\centering
%     \includegraphics[width=0.7\textwidth]{LandmarkPoints.png}
%     \label{fig:landmarkpoints}
%     \caption{}
% \end{figure}    

%%%KLT Tell some statistics, how many total number of landmarks points have you clicked? How many number of faces are annotated?
%% GDA - How long each sequence took to landmark

\section{Evaluation}
\label{sect:evaluate}
In Section \ref{sect:freeform}, we first provide a brief discussion of existing free-form registration techniques, and the three state-of-the-art techniques that we choose to evaluate our dataset. Section \ref{sect:metric} discusses our evaluation metric, and finally Section \ref{sect:results} discusses evaluation results showing the usefulness of our new dataset.

\subsection{Existing Free-Form Registration Techniques}
\label{sect:freeform}
There are a variety of different non-rigid registration techniques proposed in the literature~\cite{HUANG2008,YOU2017119,GUO2015,yang2015sparse,MA2016,yang2017robust}. For the registration of 3D face scans model-based approaches have proved popular~\cite{CHENG20173,NAIR2009}; however, these methods need a sufficiently large and representative range of training data to accurately register complex facial expressions, and require landmark points to build the underlying statistical models. If trained correctly, such techniques out-perform model-free approaches due to the prevalence of non-isometric deformations on the face, which common regularisation methods fail to cope with. Our dataset provides new extreme facial deformations as training data, complete with landmark points. Thus, they can be easily used to build a statistical model, or be used in existing model-based registration techniques.

Here, in our experiments, we are interested in registration techniques using model-free approaches, because our dataset is to be used as a challenging unseen benchmark for these techniques, motivating the research \& development to address challenging real-life datasets. We have selected two pairwise non-rigid registration techniques that use the current state-of-the-art regularisation constraints, and compare them against the baseline implementation of non-rigid ICP \cite{bouaziz2013dynamic}. We summarise these techniques below.
\begin{itemize}
\item \noindent\textit{\textbf{NICP}} \cite{bouaziz2013dynamic} is an archetypal implementation of non-rigid iterative closest point (ICP) algorithm. It iteratively computes correspondences based on a closest point criterion, and uses them to compute a series of local transformations. It requires to solve a least-square fitting problem, comprising of a fitting term with point-to-point and point-to-plane distances, and a smoothness term using global rigidity and local as-rigid-as-possible (ARAP) regularisation constraints.
\item \noindent\textit{\textbf{L12}} \cite{yang2015sparse}, similar to \cite{bouaziz2013dynamic}, incorporates pre-computed correspondences based on SHOT local shape features \cite{tam2014diffusion} and formulates an ARAP rigidity constraint as an $\ell_{1}$-norm smoothness term into the non-rigid ICP iterative pipeline. The smoothness term theoretically ensures the method is robust to noise and outliers.
\item \noindent\textit{\textbf{L11}} \cite{yang2017robust} (\textbf{L11}) extends \cite{yang2015sparse}, using both $\ell_{1}$-norm fitting and smoothness terms in an iterative framework.
\end{itemize}
None of these methods use image saliency or surface texture in the registration pipeline. \cite{bouaziz2013dynamic} does not use SHOT correspondences. For consistency, SHOT correspondences between source and target surfaces are pre-computed once and used on each method that requires them.

%The next approach incorporates pre-computed SHOT correspondences \cite{tam2014diffusion} into the non-rigid ICP pipeline with an extended ARAP rigidity constraint \cite{chen2017rigidity} In our experiment, we use a 2-ring ARAP as the default parameter. Similarly, \cite{yang2015sparse} (\textbf{L12}) also incorporates pre-computed SHOT correspondences; however, this method incorporates an $\ell\textsubscript{1}$-norm smoothness term into an iterative framework, which theoretically ensures the method is robust to noise and outliers. \cite{yang2017robust} (\textbf{L11}) extends \cite{yang2015sparse}, using $\ell\textsubscript{1}$-norm fitting and smoothness terms in an iterative framework.
%KLT \cite{chen2017rigidity} is not used in our technique, so we are not discussing it.

\subsection{Evaluation Metrics}
\label{sect:metric}
To compare the performance of these registration techniques, we first select some key frames from our funny face dataset, then apply registration techniques on them, and finally evaluate the results with annotated landmark points.

\begin{figure*}[t]
	\centering
    \includegraphics[width=0.8\linewidth]{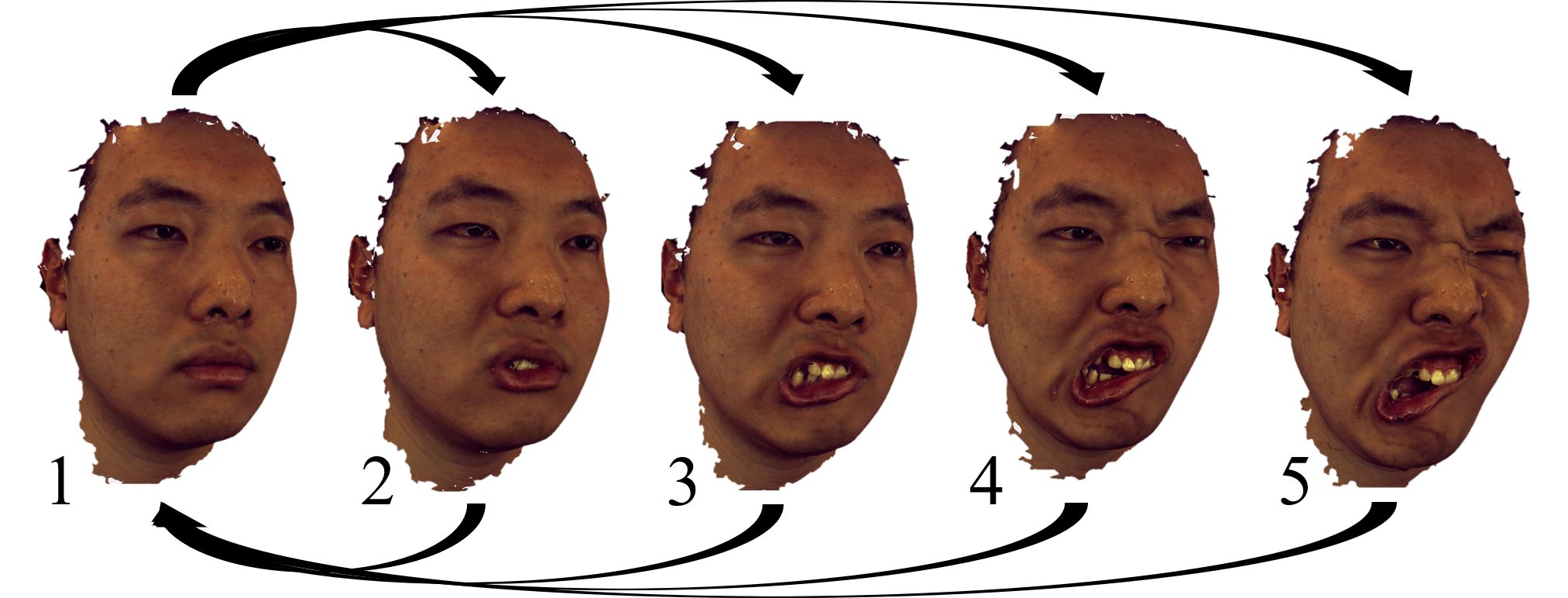}
    \caption{\label{fig:taiweiregistration} Five frames with increasing facial deformation are chosen from a funny face sequence. Frame 1 - the last frame of a neutral face. Frame 5 - the first peak frame of facial deformation. Frames 2-4 are evenly distributed between Frames 1 and 5. The arrows X$\,\to\,$Y show all pairwise registration we applied to register Frame X to Y, in our error evaluation.}
    \vspace{-0.3cm}
\end{figure*}

\textbf{Frames Selection.}
For each funny face sequence in our dataset, 5 frames of increasing level of deformation were selected. Since different sequences may have different durations, we develop the following  protocol to ensure selected frames provide sufficient and meaningful coverage of deformation levels.
The first frame is selected as the last neutral face, and the fifth is selected as the first frame with peak facial deformation. Frames 2-4 are selected with even intervals  between Frames 1 and 5. Figure \ref{fig:taiweiregistration} shows one example of these frames. For the rest of this paper, Frames 1-5 refer to these chosen representative frames.

\textbf{Pairwise registration and Error evaluation.}
The three previously mentioned pairwise non-rigid registration techniques are applied on the five representative frames of each sequence. Registration is performed in a pairwise manner (see Figure \ref{fig:taiweiregistration}) and in both forward and backward direction. The registered landmark points are found by using barycentric coordinates on the deformed source surface. Once we convert the registered points into Cartesian coordinates, We use Euclidean distance between the registered point and the annotated ground truth landmark point on the target frame as the absolute error for measure. The smaller the value, the better the accuracy of registrations.

\begin{figure}[t]
	\centering
    \includegraphics[width=\linewidth]{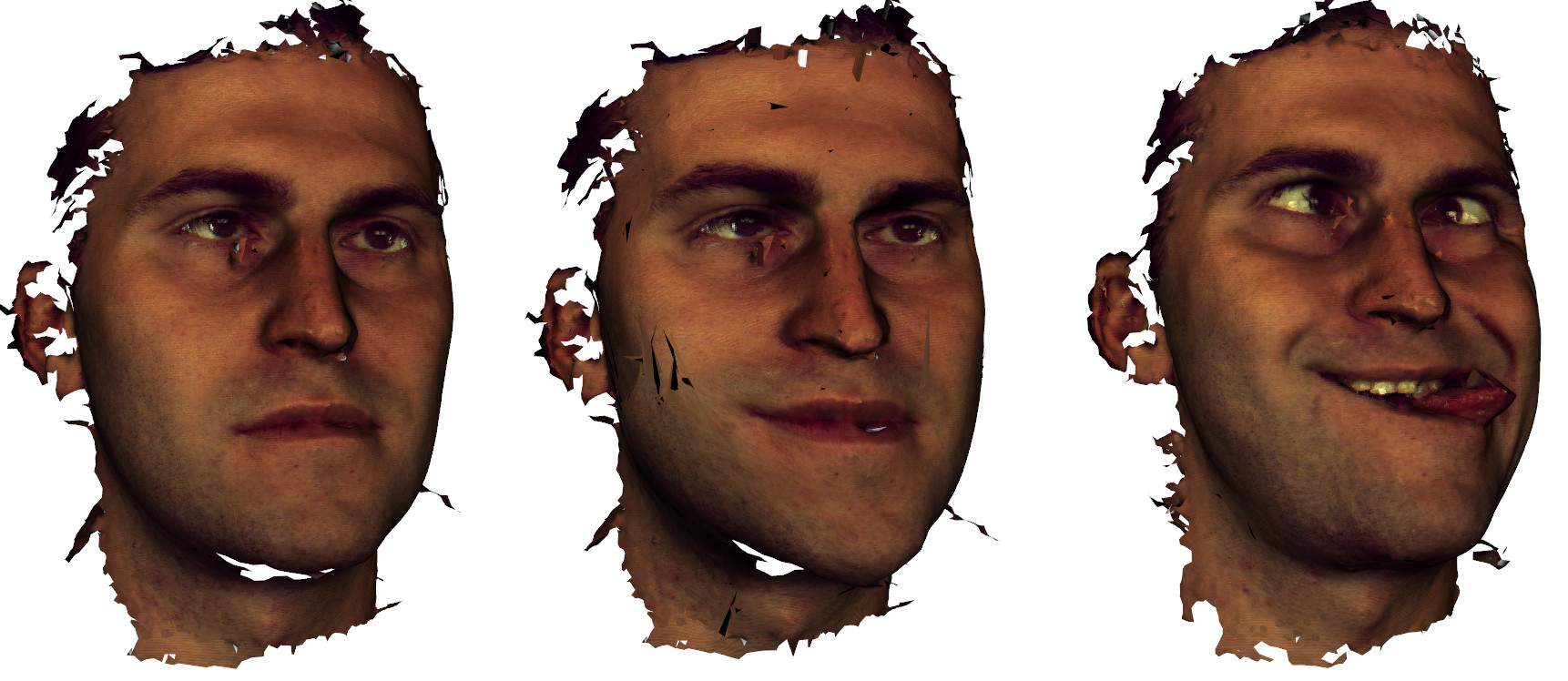}
    \caption{\label{fig:sion_comparison} Registration results of \textbf{L12}. Left: Frame 1 (source). Right: Frame 5 (target). Middle: registered results of \textbf{L12} from Frame 1 to 5. We show the texture for clarity, but texture is not used in \textbf{L12}. \textbf{L12} is able to slightly deform the mouth region, but is not able to model the large mouth movement.}
    \vspace{-0.5cm}
\end{figure}

\begin{table*}[t]
\centering
\caption{Absolute errors of the registration techniques on our challenging funny face sequences. One unit error corresponds to roughly 1.2mm.}
\label{tbl:abserrors}
\begin{tabular}{|c|cccc|cccc|cccc|}
\hline
\multirow{2}{*}{Participant} & \multicolumn{4}{c|}{L11}                                                               & \multicolumn{4}{c|}{L12}                                                               & \multicolumn{4}{c|}{NICP}                                                              \\
                             & 1$\leftrightarrow$2 & 1$\leftrightarrow$3 & 1$\leftrightarrow$4 & 1$\leftrightarrow$5 & 1$\leftrightarrow$2 & 1$\leftrightarrow$3 & 1$\leftrightarrow$4 & 1$\leftrightarrow$5 & 1$\leftrightarrow$2 & 1$\leftrightarrow$3 & 1$\leftrightarrow$4 & 1$\leftrightarrow$5 \\
\hline                             
A                            & 139                 & 245                 & 553                 & 751                 & 160                 & 257                 & 538                 & 759                 & 202                 & 407                 & 639                 & 651                 \\
B                            & 170                 & 414                 & 582                 & 1161                & 191                 & 426                 & 581                 & 1222                & 305                 & 730                 & 871                 & 1558                \\
C                            & 166                 & 245                 & 378                 & 601                 & 192                 & 260                 & 390                 & 586                 & 205                 & 397                 & 487                 & 1097                \\
D                            & 202                 & 578                 & 1040                & 1447                & 205                 & 536                 & 1029                & 1544                & 363                 & 964                 & 1016                & 1217                \\
E                            & 155                 & 252                 & 390                 & 482                 & 183                 & 275                 & 423                 & 495                 & 722                 & 480                 & 730                 & 633                 \\
F                            & 188                 & 346                 & 905                 & 1048                & 205                 & 357                 & 891                 & 1063                & 529                 & 538                 & 1110                & 1286                \\
G                            & 153                 & 287                 & 340                 & 359                 & 174                 & 297                 & 340                 & 371                 & 204                 & 375                 & 394                 & 350                 \\
H                            & 122                 & 214                 & 365                 & 345                 & 142                 & 222                 & 354                 & 349                 & 482                 & 780                 & 656                 & 770                 \\
I                            & 274                 & 476                 & 896                 & 726                 & 276                 & 493                 & 866                 & 713                 & 262                 & 653                 & 920                 & 1101                \\
J                            & 226                 & 312                 & 425                 & 521                 & 266                 & 334                 & 435                 & 530                 & 251                 & 544                 & 699                 & 872                 \\
K                            & 151                 & 363                 & 485                 & 560                 & 180                 & 389                 & 469                 & 580                 & 408                 & 462                 & 569                 & 709                 \\
L                            & 126                 & 278                 & 451                 & 604                 & 157                 & 301                 & 453                 & 601                 & 617                 & 996                 & 694                 & 854                 \\
M                            & 148                 & 279                 & 677                 & 916                 & 168                 & 284                 & 661                 & 890                 & 199                 & 344                 & 1030                & 1175                \\
N                            & 189                 & 407                 & 562                 & 523                 & 199                 & 420                 & 572                 & 512                 & 486                 & 788                 & 797                 & 866                 \\
O                            & 247                 & 439                 & 823                 & 1964                & 258                 & 452                 & 839                 & 2183                & 536                 & 693                 & 1351                & 1371                \\
P                            & 162                 & 353                 & 489                 & 740                 & 179                 & 361                 & 535                 & 774                 & 243                 & 458                 & 727                 & 1024                \\
Q                            & 219                 & 429                 & 821                 & 928                 & 246                 & 426                 & 831                 & 918                 & 362                 & 495                 & 1079                & 1048                \\
R                            & 162                 & 252                 & 451                 & 462                 & 196                 & 272                 & 429                 & 482                 & 244                 & 499                 & 689                 & 662                 \\
S                            & 202                 & 286                 & 335                 & 508                 & 211                 & 284                 & 320                 & 482                 & 363                 & 421                 & 630                 & 752                 \\
T                            & 172                 & 331                 & 394                 & 1895                & 192                 & 337                 & 390                 & 1639                & 262                 & 572                 & 644                 & 647                 \\
\hline
Avg                          & 179                 & 339                 & 568                 & 827                 & 199                 & 349                 & 567                 & 835                 & 362                 & 580                 & 787                 & 932        		 \\ 
\hline
\end{tabular}
\end{table*}

\subsection{Results, Visualisation \& Discussion}
\label{sect:results}
The results (absolute errors) of the three registration techniques on our challenging funny face dataset are shown in Table \ref{tbl:abserrors}.

From the table, there is a significant variance of errors between sequences. This reflects well the nature of our dataset, as participants were free to choose their own funny face. Some of them make very substantial deformation, whilst others are less intense. We use average values for the purpose of comparing methods. In term of error measures, \textbf{L11} and \textbf{L12} perform similarly in most of the cases. Both perform a lot better than the simple \textbf{NICP} method. It confirms that the state-of-the-art regularisation constraints have positive effects on registration results.

However, as seen in Figure \ref{fig:sion_comparison}, deformation is still a challenging task for these state-of-the-art free-form registration techniques. The deformation of the jawline is mostly captured well, and the cheek structure is approaching that of the target frame, because these areas are mostly rigidly transformed. However, the complex deformation of the mouth and tongue is not modelled well. The deformed frame also features several artefacts, in the form of black anomalies (on the cheeks). 

\begin{figure}[t]
	\centering
    \includegraphics[width=0.95\linewidth]{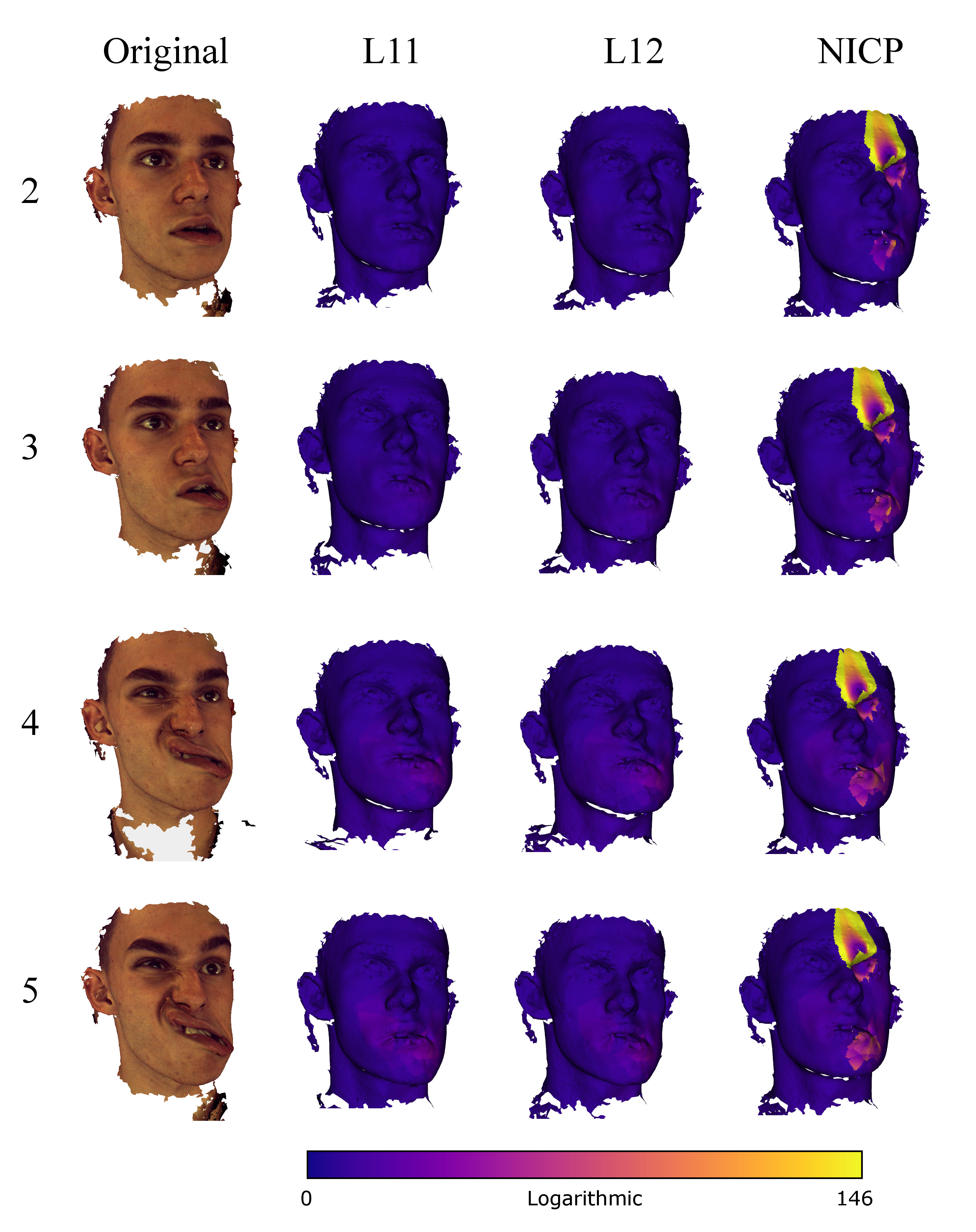}
    \caption{Visualisation of errors when registering Frame 1 to Frames 2,3,4,5: The left column shows the original output of the 3dMD system of frames 2-5 (the target surfaces). The subsequent three columns show the respective deformed source surfaces (frame 1) to the respective targets by each of the three methods, namely \textbf{L11}, \textbf{L12} and \textbf{NICP}. The target surface in Frames 2,3,4,5, as shown in separate rows, correspond to increasing degrees of facial deformation. Errors are log normalised, then mapped using plasma colourmap. This allows errors to be seen between rows despite the large range on a perceptually uniform sequential colourmap. }
    %The rest of the table shows the result of applying the specified registration technique, deforming the first frame of the sequence to nth frame. }
    \label{fig:marceli_comparison}   
\end{figure}

To confirm these numerical results, we further visualise the error distribution in Figure \ref{fig:marceli_comparison} comparing all three methods. All errors are normalised by the largest error of all sequences shown, and error values are colour coded by interpolation.

In Figure \ref{fig:marceli_comparison}, we visualize the results of all three registration techniques. The error distribution correlates well with the errors seen under visual inspection (e.g. Figure \ref{fig:sion_comparison}). When the extreme facial deformation approaches its peak, higher error values around the mouth are shown. This shows that these techniques struggle to handle such complex, elastic deformation of mouth and tongue. \textbf{NICP} also struggles with the less extreme deformations as seen in the eyebrows and forehead.

This further confirms our dataset is highly challenging, and can be used to evaluate new surface registration techniques, tackling highly elastic deformation and subtle change in topology.

%\clearpage
\section{Conclusion}
\label{sect:conclude}
In this paper, we present the problem of 4D non-rigid registration. The lack of benchmark datasets is a limiting factor in the development of this research area. We have created a new and challenging 4D facial dynamic dataset, including over 40,000 frames of 3D models from 21 participants, featuring spoken Welsh language (20 sequences landmarked) and asymmetric highly elastic deforming faces (20 sequences landmarked) annotated with 68/73 3D landmarks. Welsh spoken dataset is further annotated with level of fluency. Furthermore, we also evaluate the performance of existing registration techniques on our challenging funny face dataset, using respective error measure and supported by visualisation. We intend to release the dataset to the community to promote related research.

%%%YKL Commented out for anonymous review
%\section{Acknowledgement}
%This work was generously supported by the Coleg Cymraeg Cenedlaethol Strategic Development Fund and the SPIN Paid Internship Employment Fund.

\bibliographystyle{ieeetran}
\bibliography{egbib}
\end{document}